\documentclass[hidelinks]{isprs}
\usepackage{subfigure}
\usepackage{import}
\usepackage{setspace}
\usepackage{geometry} 
\usepackage{epstopdf}
\usepackage[labelsep=period]{caption}  

\usepackage{hyperref}
\usepackage{graphicx}
\usepackage[export]{adjustbox}
\usepackage{amsmath,amsfonts,amssymb}
\usepackage[acronyms]{glossaries}
\usepackage[sort]{natbib} 
	\renewcommand{\refname}{REFERENCES} 
\usepackage{color, soul}
\usepackage{diagbox}	
\usepackage{enumitem}
\usepackage{gensymb}
\usepackage[nameinlink, capitalize, noabbrev]{cleveref}
\usepackage{hhline}
\usepackage{inputenc}

\usepackage[percent]{overpic}

\usepackage{xspace}

\newcommand{\eg}{e.g.\ }

\newcommand{\m}{\,m\xspace}

\newcommand{\percent}{\,\%\xspace}

\begin{document}

\title{Automatic Co-Registration of Aerial Imagery and Untextured Model Data utilizing Average Shading Gradients}

\author{
 S. Schmitz\textsuperscript{b,a}, M. Weinmann\textsuperscript{b}, B. Ruf\textsuperscript{a,b}
}

\address{
\textsuperscript{a}Fraunhofer IOSB, Video Exploitation Systems, Karlsruhe, Germany -\\ \{sylvia.schmitz, boitumelo.ruf\}@iosb.fraunhofer.de\\
\textsuperscript{b}Institute of Photogrammetry and Remote Sensing, Karlsruhe Institute of Technology, \\Karlsruhe, Germany - \{martin.weinmann, boitumelo.ruf\}@kit.edu
}


\commission{II, }{I} 
\workinggroup{II/1} 
\icwg{}   


\newacronym{ASG}{ASG}{Average Shading Gradients}

\newacronym[plural={CNNs}, longplural={convolutional neural networks}]{CNN}{CNN}{convolutional neural network}
\newacronym{COTS}{COTS}{commercial off-the-shelf}
\newacronym{CT}{CT}{Census Transform}

\newacronym{DLT}{DLT}{Direct Linear Transformation}

\newacronym{EPnP}{EPnP}{Efficient Perspective-n-Point}

\newacronym{GPGPU}{GPGPU}{general purpose computation on a GPU}
\newacronym{GPS}{GPS}{Global Positioning System}

\newacronym{ICP}{ICP}{Iterative Closest Point}
\newacronym{IMU}{IMU}{Inertial Measurement Unit}
\newacronym{INS}{INS}{Inertial Navigation System}

\newacronym{LIDAR}{LiDAR}{light detection and ranging}

\newacronym[shortplural={MRFs}, longplural={Markov Random Fields}]{MRF}{MRF}{Markov Random Field}
\newacronym{MVS}{MVS}{Multi-View Stereo}

\newacronym{NCC}{NCC}{normalized cross correlation}

\newacronym{PCL}{PCL}{Point Cloud Library}

\newacronym{RANSAC}{RANSAC}{Random Sampling Consensus}
\newacronym[shortplural={ROIs}, longplural={regions of interest}]{ROI}{RoI}{region of interest}

\newacronym{SAD}{SAD}{sum of absolute differences}
\newacronym{SFM}{SfM}{Structure-from-Motion}
\newacronym{SGM}{SGM}{Semi-Global Matching}

\newacronym[shortplural={UAVs}, longplural={unmanned aerial vehicles}]{UAV}{UAV}{unmanned aerial vehicle}

\newacronym{WTA}{WTA}{winner-takes-it-all}

\abstract{
The comparison of current image data with existing 3D model data of a scene provides an efficient method to keep models up to date. In order to transfer information between 2D and 3D data, a preliminary co-registration is necessary. In this paper, we present a concept to automatically co-register aerial imagery and untextured 3D model data. To refine a given initial camera pose, our algorithm computes dense correspondence fields using SIFT flow between gradient representations of the model and camera image, from which 2D-3D correspondences are obtained. These correspondences are then used in an iterative optimization scheme to refine the initial camera pose by minimizing the reprojection error. Since it is assumed that the model does not contain texture information, our algorithm is built up on an existing method based on \gls*{ASG} to generate gradient images based on raw geometry information only. We apply our algorithm for the co-registering of aerial photographs to an untextured, noisy mesh model. We have investigated different magnitudes of input error and show that the proposed approach can reduce the final reprojection error to a minimum of 1.27\,$\pm$0.54\, pixels, which is less than 10\percent of its initial value. Furthermore, our evaluation shows that our approach outperforms the accuracy of a standard \gls*{ICP} implementation.

%
\ \\ 
}

\maketitle

\glsresetall 

\section{INTRODUCTION}
\label{sec:intro}
\sloppy
Due to technological advancements in the field of sensor technology and algorithm development for the acquisition, generation and processing of 3D data, the availability and use of 3D models has risen significantly. The advantages of 3D visualization are applied successfully in applications such as urban navigation and planning, ecological development or security surveillance. City administrations, for example, already maintain city models which are used advantageously in spatial and urban planning. A prerequisite for a reasonable use is that these models match reality as closely as possible. However, the acquisition and provision of large-scale 3D models with a high level of detail is still very expensive and time-consuming. Consequently, such models are typically only generated every few years. In order to compensate for the time gap between the acquisition and dissemination of publicly accessible 3D models, we work on augmenting the model data with up-to-date aerial imagery and perform a change detection to update the existing model.

\begin{figure}
	\centering
	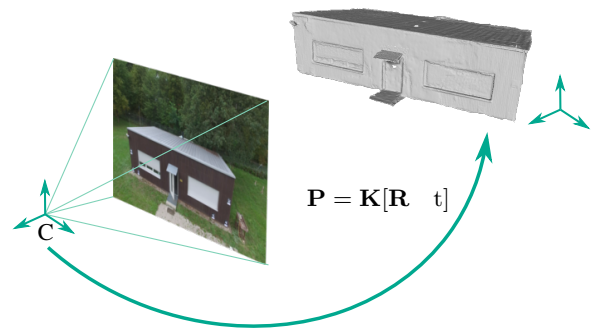
	\caption{Registration task: Estimation of camera parameters, which describe the relative pose between image and model.}
	\label{fig:registration}
\end{figure}
In order to enable the transfer of information between aerial imagery and a 3D model, a preliminary co-registration is necessary. This task can be formulated as the estimation of camera parameters which describe the relative position and orientation between image and model as illustrated in \cref{fig:registration}. Given point correspondences between image and model, this is a well-known task for which numerous efficient methods have already been developed (for instance proposed by \cite{Lepetit2009}, \cite{penate2013exhaustive} or \cite{gao2003complete}). If there is no information available regarding corresponding points, the main challenge is to identify discriminative points which can be used to estimate the correspondences between the model and the image. If the model data is textured, these correspondences can be identified by exploiting image features (\eg SIFT, SURF or ORB). However, texture of the model cannot always be relied on, as it might be outdated, of different modality, or simply missing. For this reason, we present a concept to automatically co-register aerial imagery and untextured 2.5D or 3D models, which is working with raw geometry information only. 

Our algorithm is based on the concept for automatic registration of images to untextured geometry, which has been proposed by \cite{Plotz2017asg}. Assuming an initial camera pose, a given untextured 2.5D or 3D model and a camera image, our algorithm estimates the intrinsic and extrinsic camera parameters based on 2D-3D correspondences between pixels in the input image and object points in the model. In order to detect reliable correspondences, image features of the input image are compared with features extracted on rendered views of the 3D model. A basic component of many proven feature descriptors consists of intensity gradients, which we also use in our algorithm to describe and compare image features. Since we deal with an untextured model, the calculation of texture gradients on rendered images generated from the model is not possible. To this end, we rely only on gradients of shadings by applying the so-called \gls*{ASG} proposed by \cite{Plotz2017asg}. This is a rendering technique in which observable shading gradients are averaged over all possible lighting directions of the 3D scene under the assumption of a local illumination model.

This paper is organized as follows. After presenting related work that focuses on image-to-model co-registration, we describe our algorithm in Section 3. There, we explain the generation of gradient images as well as the correspondence search and pose estimation. In Section 4, we present and discuss our dataset and the results we achieved with our approach. Finally, we provide concluding remarks and suggestions for future work in Section 5.

\section{RELATED WORK}%
\label{sec:related_work}
\sloppy

Using 2D-3D correspondences to estimate the relative camera pose of a single query image with respect to a given 3D model is a widely applied approach. Similar to the concept of this paper, \cite{Russell2011}, \cite{aubry2014}, \cite{Irschara2009} and \cite{Sibbing2013} determine the necessary correspondences by comparing the query image to rendered views of the model. The approach of \cite{Irschara2009} is based on a mapping of SIFT features extracted from the query image to SIFT features extracted from images of a database. The database used consists of already registered images as well as images of the 3D model, which were generated from different viewpoints. In a similar manner, \cite{Russell2011} and \cite{aubry2014} present approaches that generate images from a model to register paintings. \cite{Russell2011} combine the approach of \cite{Irschara2009} with a matching of GIST descriptors \citep{Oliva2001} to find a rough camera pose. The co-registration is then improved by matching the contours in the painting to the view-dependent contours of the model. \cite{aubry2014}, on the other hand, use rendered images from a model to identify features that are reliably recognizable in any 2D representation and are used for matching. The features used in the approaches mentioned are based on texture information contained in the 3D model. In contrast, this paper assumes an untextured model. In addition, \cite{Russell2011} and \cite{aubry2014} assume that the camera is located near the ground, while in this work we focus on the co-registration of aerial imagery.

Likewise, the task of co-registering aerial photographs to a 3D model is addressed by \cite{vasile2006}, \cite{frueh2004} and \cite{mastin2009}. In \citep{frueh2004}, the goal is to transfer texture information from aerial photographs to a city model. The required registration of the images is based on an assignment of projected 3D lines of the model to 2D lines, which are extracted in the aerial photograph. Using a given camera pose with an accuracy comparable to that of a \gls*{GPS} and \gls*{INS} system, an exhaustive search is performed using extrinsic and intrinsic camera parameters. \cite{vasile2006} present a similar approach. Pseudo-intensity images are generated using shadows of \gls*{LIDAR} data which are compared to the 2D aerial image by correlation. Similar to \citep{frueh2004}, the complete camera pose is then determined by an exhaustive search with \gls*{GPS} information providing an initial estimate. Both methods lead to accurate registrations, but are very time-consuming. The goal to transfer information from images to a model is also pursued by \cite{mastin2009}. In their approach, aerial images are registered with respect to a \gls*{LIDAR} point cloud in order to generate a photorealistic 3D model. The registration process based on Mutual Information determines camera parameters that maximize the transinformation between the distribution of image features and projected \gls*{LIDAR} features. Used \gls*{LIDAR} features arise from elevations and probability of detection, whereas image features result from illumination intensities. The camera parameters that maximize the transinformation are determined using simplex methods. The use of OpenGL and graphics hardware in the optimization process leads to significantly shorter registration times compared to the methods of \cite{frueh2004} and \cite{vasile2006}. The three methods co-register aerial photographs with respect to a \gls*{LIDAR} point cloud. In contrast, the aim of our work is to register an aerial photograph to a meshed 3D model, independent from the image sensor technology used to generate the model.

Lately, learning-based approaches are in particular attracting a great deal of attention. Solutions based on \glspl*{CNN}, proposed by \cite{Brachmann2018}, \cite {Kendall2015} and \cite{Hou2018}, are presented to solve the task of 2D/3D co-registration. In order to determine the camera pose in a given 3D environment using an RGB image, \cite{Brachmann2018} propose a \gls*{CNN} that predicts 2D-3D correspondences. Hypotheses regarding the camera parameters are determined from four correspondences each. A second \gls*{CNN} is used to determine the most probable camera pose. \cite{Kendall2015} describe the training of a \gls*{CNN} that directly estimates extrinsic camera parameters based on an image. The required training data are generated by a \gls*{SFM} process and used for the fine tuning of pre-trained models. \cite{Hou2018} also estimate the pose directly via a \gls*{CNN}, proposing a new loss function based on Riemann's geometry. In contrast to learning-based approaches, this paper presents a method that does not require complex training procedures or a large amount of training data. Updating and adapting the method to other conditions, such as prior knowledge of the camera pose, is easily possible without having to train new models.

\section{METHODOLOGY}%
\label{sec:methodology}
\sloppy
\begin{figure*}
	\centering
	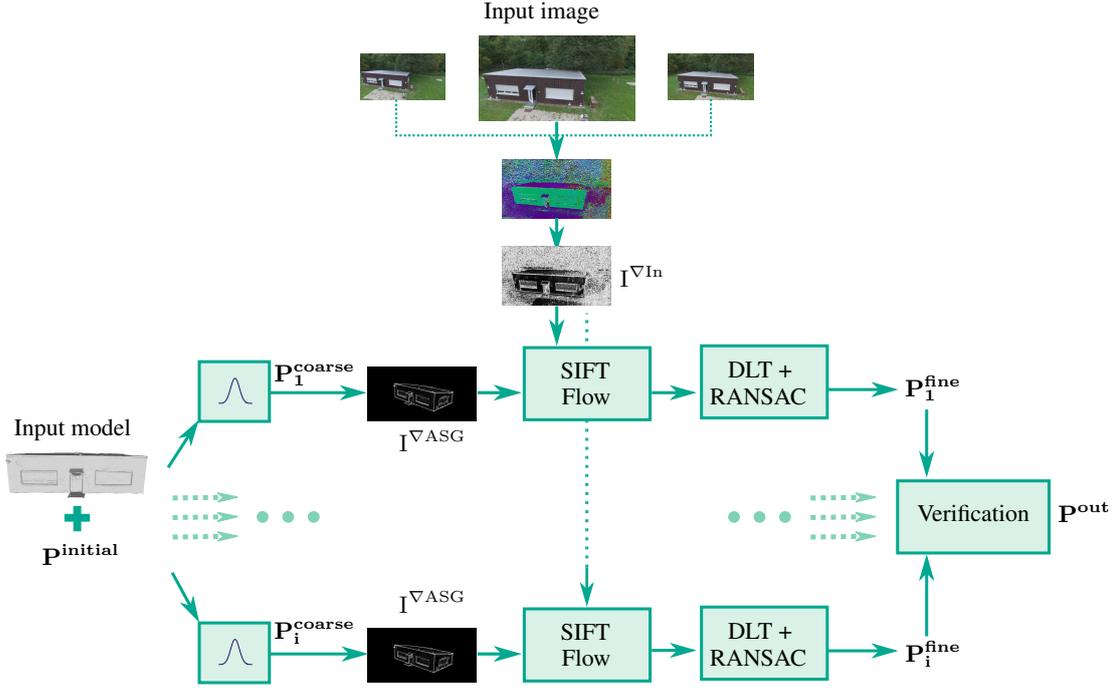
	\caption{Overview of the processing pipeline of the co-registration algorithm: Gradient images rendered from the model are matched to a gradient image of the photograph using SIFT flow. The correspondences determined are used to estimate the relative pose based on a DLT. The final pose is selected in the verification step.}
	\label{fig:procedure}
\end{figure*}

The overall procedure to estimate the relative camera pose between an untextured model and a camera image (referred to as query image) is depicted in \cref{fig:procedure}. Given an untextured 3D model, an initial camera pose $\mathrm{P}^{\mathrm{initial}}$ and a depth image generated by \gls*{SFM} from the camera images, our algorithm utilizes gradient representations of the input model and query image to extract features which are used in an iterative optimization scheme to refine the initial camera pose. Using coarse poses $\mathrm{P}^{\mathrm{coarse}}_i$ close to the initial camera pose, we render the model with \gls*{ASG}, yielding multiple gradient images of the input model. The query image is also transformed into a gradient representation. Thus, at the end of the first step, we have a set of gradient images $\mathrm{I}^{\mathrm{\nabla ASG}}_i$ extracted from the 3D model associated with coarse camera poses $\mathrm{P}^{\mathrm{coarse}}_i$ and one gradient image $\mathrm{I}^{\mathrm{\nabla In}}$ corresponding to the query image. In the second stage, we compute a dense correspondence field between each $\mathrm{I}^{\mathrm{\nabla ASG}}_i$ and $\mathrm{I}^{\mathrm{\nabla In}}$ using the SIFT flow algorithm \citep{Liu2011sift}. %
With these 2D-3D correspondences, we employ a \gls*{DLT} \citep{Hartley2003} within a RANSAC loop to iteratively refine each coarse camera pose $\mathrm{P}^{\mathrm{coarse}}_i$ to receive $\mathrm{P}^{\mathrm{fine}}_i$. Given the set of refined camera poses, a final verification step selects the pose $\mathrm{P}^{\mathrm{fine}}_i$ with the smallest reprojection error as the output pose $\mathrm{P}^{\mathrm{out}}$.

\subsection{Gradient Representations}

\begin{figure}
	\centering
	\resizebox{0.5\textwidth}{!}{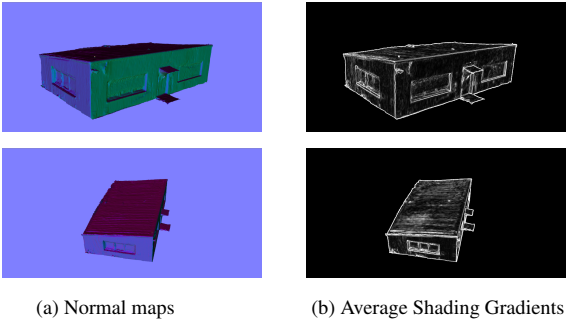}
	\caption{Gradient images created using Average Shading Gradients. (a) Normal maps of the model, (b) Resulting gradient images.}
	\label{fig:ASG}
\end{figure}

The first step of the registration process is the computation of a set of gradient images, which represents the model from different perspectives. For this purpose, we perpetuate the initial extrinsic camera parameters with Gaussian noise to generate multiple views on the model distributed around the provided camera pose $\mathrm{P}^{\mathrm{initial}}$. Gradient images are generated from the model using the generated coarse camera poses $\mathrm{P}^{\mathrm{coarse}}_i$. Since the model does not contain any color or gray value information, gradients can result solely from shadings due to specific lighting of the scene. Hence, we use \gls*{ASG} in the same manner as in \citep{Plotz2017asg}. \ \\
A gradient image can be described by convolving the image matrix and derivative filters in the x and y directions: \\
\begin{align}
	 ||\nabla \mathrm{I} || = \sqrt{\left( \mathrm{h_x} *\mathrm{I} \right) ^2 + \left( \mathrm{h_y} *\mathrm{I} \right) ^2},
\end{align}\\
with $\mathrm{I}$ being the image and $\mathrm{h_x}$, $\mathrm{h_y}$ indicating the derivative filters. Under the assumption of a Lambertian illumination model and the use of a point light source, the intensities of a rendered image $\mathrm{I}$ can be described by: \\
\begin{align}
	 \mathrm{I} = \max (0, - \mathrm {n}^\top \mathrm {l}). 
\end{align} \\
In this formulation, the normal direction is expressed by the vector $\mathrm{n}$ and the direction of light by the vector $\mathrm{l}$. With predefined light direction $\mathrm{l}$, the combination of the above equations allows the calculation of a rendered image in gradient representation. However, the assumption of a fixed light direction has considerable disadvantages. Discontinuities in the normal map, which cause shadings under an assumed light direction and thus high values in the gradient image, do not lead to observable gradients under the assumption of a different light direction. The approach of \gls*{ASG} counteracts this undesirable behavior. The observed gradient strengths are averaged over all feasible light directions $\mathrm{l}$ along the unit sphere $\mathcal{S}$. The corresponding mathematical description is:\\
\begin{align}
	\overline{||\nabla \mathrm{I}||} = \int _{\mathcal{S}} \Big [\left( \mathrm{h_x} *\max (0, - \mathrm {n}^\top \mathrm {l}) \right) ^2 \nonumber \\ +\, \left( \mathrm{h_y} *\max (0, - \mathrm {n}^\top \mathrm {l}) \right) ^2 \Big ]^{\frac{1}{2}}\;\text {d}\mathrm {l}.
\end{align}\\
The exact calculation of $\overline{||\nabla \mathrm{I}||}$ is very computation-intensive due to the complex integrand. Therefore, approximations are proposed by \cite{Plotz2017asg} that allow the estimation of $\overline{||\nabla \mathrm{I}||}$ in closed form:\\
\begin{align}
	\label{eq:ASG}
	\overline{||\nabla \mathrm{I}||} &\approx \dfrac{1}{2} \int _{\mathcal{S}} \Big [\left( \mathrm{h_x} * \mathrm {n}^\top \mathrm {l} \right) ^2 \nonumber \, +\, \left( \mathrm{h_y} * \mathrm {n}^\top \mathrm {l} \right) ^2 \Big ]^{\frac{1}{2}}\;\text {d}\mathrm {l}
	\\&	\le \frac{1}{2} \sqrt{ \int _{\mathcal{S}} \left( \mathrm{h_x} *(\mathrm {n}^\top \mathrm {l}) \right) ^2 + \left( \mathrm{h_y} *(\mathrm {n}^\top \mathrm {l}) \right) ^2 \;\text {d}\mathrm {l}} \nonumber \\&= \frac{1}{2} \sqrt{ \int _{\mathcal{S}} \left( (\mathrm{h_x} *\mathrm {n})^\top \mathrm {l}\right) ^2 \;\text {d}\mathrm {l}+ \int _{\mathcal{S}}\left( (\mathrm{h_y} *\mathrm {n})^\top \mathrm {l}\right) ^2 \;\text {d}\mathrm {l}} \nonumber \\&= \sqrt{\frac{\pi }{3}} \sqrt{ \sum _{i=1}^3 (\mathrm{h_x} *\mathrm {n}_i)^2 + (\mathrm{h_y} *\mathrm {n}_i)^2 }.
\end{align}
The designation $\mathrm{n}_i$ indicates the x-, y- and z-component of the normal map. For the proof of the equality of the expressions in \cref{eq:ASG}, we refer to \citep{Plotz2017asg}. The represented form enables an efficient calculation of the desired gradient image, which is based exclusively on the convolution of the normal map with derivation filters. \cref{fig:ASG} shows result images, which can be derived by means of \gls*{ASG}.

We use the presented approximation of \gls*{ASG} to render one gradient image for each generated coarse pose $\mathrm{P}^{\mathrm{coarse}}_i$. Features of these images are to be matched with features of the query image that is to be registered. In order to get comparable features, the input image also needs to be converted to a representation that contains gradients induced by shadings. Applying discrete derivative operators on the single camera image, as proposed by \cite{Plotz2017asg}, results in gradients which are related to \gls*{ASG} computed on the model to a certain degree. To obtain even more comparable gradients, we include additional camera images captured with a small spatial offset to the query image. The added images allow to compute depth information and normal vectors for the query image by means of \gls*{SFM}. Consequently, we can utilize \gls*{ASG} to transform the input image into a representation $\mathrm{I}^{\mathrm{\nabla In}}$ equivalent to the gradient representations $\mathrm{I}^{\mathrm{\nabla ASG}}_i$ of the renderings.

\subsection{Feature Matching}

In order to match the rendered images to the query image, we use the SIFT flow algorithm \citep{Liu2011sift} computing dense flow fields between each $\mathrm{I}^{\mathrm{\nabla ASG}}_i$ and $\mathrm{I}^{\mathrm{\nabla In}}$. This algorithm works similar to the optical flow method, determining a pixel-wise shift between two images. Instead of computing correspondences by individual pixel intensities, the matching in the SIFT flow algorithm is based on the comparison of SIFT descriptors calculated for each pixel. In order to prevent the assignment of pixels from regions of homogeneous structure in the input image to empty regions in the rendering, only pixels located in textured regions are included in the calculation of the flow. 

To compute correspondences, the flow vectors are first determined from the query gradient image $\mathrm{I}^{\mathrm{\nabla In}}$ to each rendered gradient image $\mathrm{I}^{\mathrm{\nabla ASG}}_i$, then from each rendered image $\mathrm{I}^{\mathrm{\nabla ASG}}_i$ to the query gradient image $\mathrm{I}^{\mathrm{\nabla In}}$. Pixel pairs connected by two opposite flow vectors are recorded as corresponding pair. The determined point correspondences are converted into 2D-3D correspondences between the input image and the model. The 3D model points are reconstructed from the appropriate points of the rendered images by considering the associated coarse camera poses $\mathrm{P}^{\mathrm{coarse}}_i$.

\subsection{Pose Estimation}

Given the determined 2D-3D correspondences, we improve successively each coarse pose $\mathrm{P}^{\mathrm{coarse}}_i$. Assuming that enough correct correspondences have been detected, a RANSAC scheme can be used to reliably determine the relative camera pose between the query image and the model. Within the inner RANSAC loop, six 2D-3D point pairs are randomly selected from the available correspondences. From these, intrinsic and extrinsic camera parameters are determined by applying the \gls*{DLT} algorithm \citep{Hartley2003}. Subsequently, we check how many of the given correspondences support the determined camera pose and form a consensus set by computing the reprojection error for each correspondence. The camera pose corresponding to the largest consensus set represents the result and thus the improved camera pose $\mathrm{P}^{\mathrm{fine}}_i$. Empirically, we have found that a good termination criterion is the computation of a consensus set that contains at least 65\percent of all correspondences. If this criterion is not met, the calculation is aborted after a maximum number of 500 iterations. 

\subsection{Plausibility Check}

The final step of our automatic co-registration process selects the most appropriate camera pose from all the refined poses $\mathrm{P}^{\mathrm{fine}}_i$. Since a correct registration is not possible in every case, we also use this step to decide on the success of the registration. For this purpose, the mutual reprojection error is calculated in pairs between all refined pose estimates. This is defined by the following equation:\\
\begin{align} 
	\label{eq:Reprojektionsfehler}
	\delta (\mathcal {P}, \mathcal {P}')= & {} \frac{1}{2} \left( \frac{1}{ \left| \mathcal {V}\right| } 	\sum _ {\mathbf {x} \in \mathcal {V}} ||\mathcal {P}(\mathbf {x}) - \mathcal {P}'(\mathbf {x}) ||_2 		\right. \nonumber \\&\left. +\, \frac{1}{\left| \mathcal {V}'\right| } \sum _ {\mathbf {x} \in 				\mathcal {V}'} ||\mathcal {P}(\mathbf {x}) - \mathcal {P}'(\mathbf {x}) ||_2 \right) . 
\end{align}\\
$\mathcal {P}$ and $\mathcal {P}'$ denote poses which project the model coordinates into an image plane; $\mathcal {V}$ and $\mathcal {V}'$ denote the set of visible points in the image area. Thus the error measure describes the mean Euclidean distance between visible pixels. If the error for two poses is below a threshold value, which is fixed to 5\percent of the longest image dimension, the considered poses are considered compatible. An undirected graph is defined by the compatibility relationship of all poses to each other. The nodes of the graph each represent a camera pose. The edges of the graph represent the compatibility between camera poses. By means of a depth-first search, the largest connected component of the graph is determined. If this contains more than three nodes, the registration process is considered successful. As the final pose  $\mathrm{P}^{\mathrm{out}}$ we select the pose from the largest connected component of the graph, for which the largest consensus set was reached in the previous RANSAC scheme.

\section{EXPERIMENTS}
\label{sec:eval}
\begin{figure}
	\begin{minipage}[b]{0.5\textwidth}
	\centering
	\subfigure[Input images]{\label{fig:input images} \resizebox{0.9\textwidth}{!}{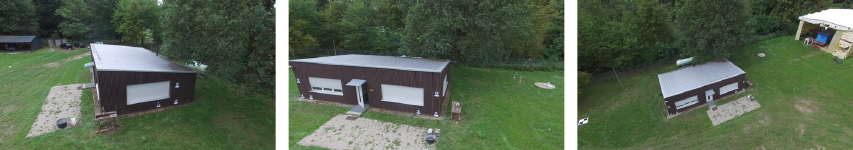}}
	\end{minipage}
	\begin{minipage}[b]{0.5\textwidth}
	\centering
	\subfigure[Point cloud]{\label{fig:b}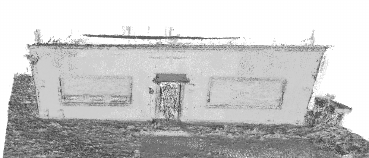}
\subfigure[Mesh model]{\label{fig:c}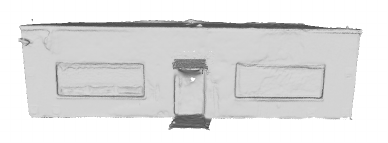}
	\end{minipage}
	\caption{(a) Example images of our dataset. The point cloud generated from the images can be seen in (b).(c) Resulting mesh model.}
\end{figure}

\sloppy
The quantitative evaluation of the registration procedure is based on a dataset comprising 164 aerial photographs captured with a DJI Phantom 3 Professional. These depict a free-standing building from three sides at three different heights (2\m, 8\m and 15\m). Background objects, such as trees or people, are also depicted and may differ between the images. \cref{fig:input images} shows a selection of the aerial photographs used. Based on all given images, a 3D point cloud is created using the \gls*{SFM} pipeline COLMAP \citep{Schoenberger2016sfm, Schoenberger2016mvs} (\cref{fig:b}). From the relevant points, i.e. those representing the building, a surface model was generated by a triangular meshing. The Poisson Surface Reconstruction method \citep{Kazhdan2006poisson} provided in COLMAP, was used for this purpose. A view of the mesh model is shown in \cref{fig:c}. The resulting model shows some errors due to the limited accuracy of the point cloud. 

Aerial photographs are usually taken using \gls*{UAV} systems equipped with \gls*{GPS} and \gls*{IMU} sensors. This allows to derive a camera pose for each image, which can then be used to initialize the registration process. Since the dataset used does not contain any sensor data, we simulate the initial camera poses $\mathrm{P}^{\mathrm{in}}$ by applying additive white Gaussian noise to the translation and rotation parameters of the ground truth poses $\mathrm{P}^{\mathrm{GT}}$. The ground truth poses are derived from the \gls*{SFM} process on which the 3D reconstruction is based. 

For our experiments, we have selected 50 test images of the dataset associated with initial camera poses and scaled them to a size of 505\,$\times$\,275 pixels. When selecting the images, we took care to cover as wide a range of perspectives as possible. For each registration process of our tests, we generated and improved 15 coarse poses and automatically choose the best refined pose.

\subsection{Accuracy and Success Rate}

We evaluated the presented co-registration algorithm (hereinafter referred to as \gls*{ASG} approach) with respect to the accuracy of the estimated camera poses and the success rate of the registration. Using the \gls*{ASG} approach, all eleven degrees of freedom of the desired camera pose can be determined. However, in many applications, intrinsic camera parameters are given by a preliminary sensor calibration. Under these conditions, an \gls*{EPnP} method according to \citep{Lepetit2009} can be used instead of the \gls*{DLT} to estimate the camera pose from 2D-3D correspondences. The performance of the presented co-registration algorithm was evaluated for both initial conditions (intrinsic parameters known and unknown) using 50 selected images. During our tests, we successively increased the mean initial error, which represents the average displacement of the initial poses $\mathrm{P}^{\mathrm{in}}$ in relation to the true poses $\mathrm{P}^{\mathrm{GT}}$. We quantify initial errors by the mutual reprojection error as defined in \cref{eq:Reprojektionsfehler}.

\begin{figure}
	\centering
	\resizebox{0.5\textwidth}{!}{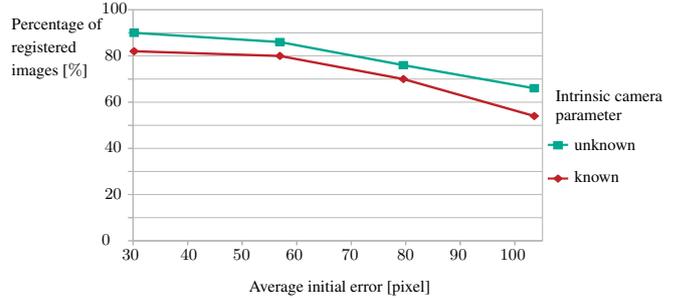}
	\caption{The rate of success for known and unknown intrinsic camera parameters.}
	\label{fig:sucessrate}%
\end{figure}
First we have examined how many images are rated as successfully co-registered by the automatic registration process. \cref{fig:sucessrate} presents the corresponding results for the assumption of known and unknown intrinsic camera parameters. The results show that, under the condition of unknown intrinsic parameters, more images are accepted as successfully registered. This is due to the fact that the camera pose can be adapted more flexibly to the found correspondences due to the higher degree of freedom. However, as shown in the following section, this is at the expense of the accuracy of the estimate. It can also be seen that, as the initial error increases, the registration task becomes more demanding and some of the images are not registered. With low initial errors of about 30 pixels, 90\percent and 82\percent  of the images are accepted within the automatic verification, whereas a medium error of about 100 pixels only results in 66\percent and 54\percent, respectively.

A second aspect that was evaluated is the improvement of the initial errored poses by the proposed co-registration. Again, we use the mutual reprojection error (\cref{eq:Reprojektionsfehler}), to evaluate the pose estimation. To summarize the distribution of the errors, the mean value, the standard deviation and the range of the reprojection errors are stated. The resulting error statistics for known and unknown intrinsic camera parameters are shown in \cref{tab:accuracy} for comparison.

\begin{table}
	\centering
	\begin{tabular}{llll}
		\hline
		\textbf{Initial error} & \textbf{Intrinsics} & \textbf{Accuracy} & \textbf{Error range}\\
		\hline \hline
		$30.19$ & unknown & $3.10 \pm 2.01$ & $1.22 - 11.93$ \\
		& known & $1.27 \pm 0.54$ & $0.57 - 2.20$\\
		\hline
		$56.90$ & unknown & $4.10 \pm 3.31$ & $1.32 - 15.75$\\
		& known & $1.54 \pm 1.37$ & $0.28 - 8.40$ \\
		\hline
		$79.53$& unknown & $4.39 \pm 3.26$ & $1.49 - 16.85$\\
		& known & $1.81 \pm 2.13$ & $0.41 - 12.86$ \\
		\hline
		$103.83$& unknown & $3.70 \pm 2.50$ & $1.87 - 12.50$ \\
		& known & $2.51 \pm 1.71$ & $0.85 - 6.50$ \\
		\hline		
	\end{tabular}
	\caption{This table holds the accuracy of our co-registration approach for various mean initial errors under known and unknown intrinsic camera parameters. The accuracy is given by the mean value and the standard deviation of the reprojection errors of registered images.}
\label{tab:accuracy}
\end{table}

The following observations are made: The maximum reprojection error with respect to all registered images is 16.85 pixels. This corresponds to about 3.3\percent of the largest image dimension. Even with a small initial error of about 30 pixels, this is equal to a reduction of the error by almost 50\percent. From this, it can be concluded that automatic verification only accepts estimates that are more accurate than the given incorrect pose $\mathrm{P}^{\mathrm{in}}$. Furthermore, it is shown that the average reprojection error can be reduced to up to 3.1 pixels when estimating all camera parameters. If the intrinsic parameters are already given, an average value of 1.3 pixels can be even achieved. Even with high initial errors of more than 100 pixels, good pose estimates with an average reprojection error of 3.7 or 2.5 pixels are achieved by our co-registration process. This corresponds to an improvement to 2.63\percent or 1.07\percent of the mean initial error. The large differences in the reprojection errors between the various test images, which can be seen in the large error ranges, are noticeable. The high standard deviations, which are in the same order of magnitude as the mean value itself, also indicate a strong scatter of the errors. Certain viewing perspectives of the scene are therefore more difficult to estimate than others. The examination of individual results shows: If different sides of the building are depicted on a query image, a more precise registration can usually be made than for images in which mainly one side of the building is visible. This is due to the influence of the spatial constellation of the correspondence points on the accuracy of pose estimation methods. Thus the equation system of the \gls*{DLT} gets singular or numerically unstable for a planar point set. The \gls*{EPnP} method explicitly distinguishes the planar point configuration from the non-planar case, resulting in increasing inaccuracy in the intermediate near-planar case.

\begin{figure*}
	\centering
	\resizebox{\textwidth}{!}{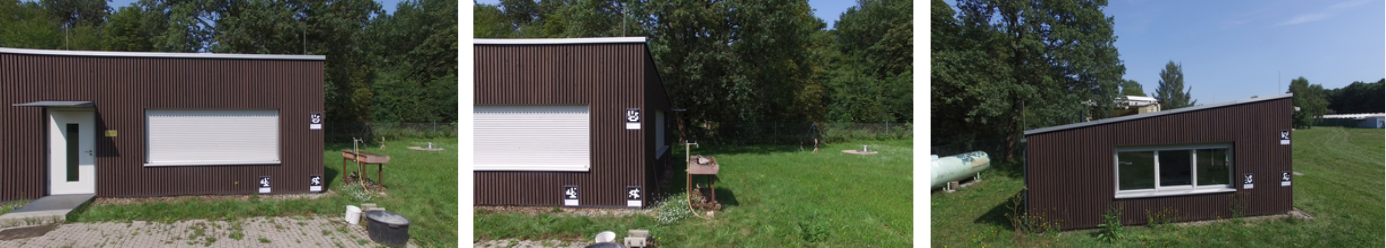}
	\caption{Images that cannot be registered by our algorithm.}
	\label{fig:negativ}
\end{figure*}

\cref{fig:negativ} shows sample images for which our registration algorithm fails. It can be seen that these are images where all visible points of the building have a similar distance to the camera.

In summary:
\begin{itemize}
	\item With the automatic verification, on the test data our registration approach only provides camera poses that are more accurate than the initial camera poses.
	\item The average reprojection error of the camera poses is improved by the registration process up to 1.07\percent of the average initial error.
	\item The achieved accuracies increase only slightly for high initial errors.
	\item With known intrinsic camera parameters, a more precise estimation of camera poses is possible.
\end{itemize}

\subsection{Comparison to an ICP Implementation}
In order to rate our results obtained, we compared the accuracy and success rate of our approach to a standard \gls*{ICP} implementation. Within the evaluation, we used the implementation of the software library \gls*{PCL} \citep{Rusu2011}. In order to use the \gls*{ICP} algorithm to estimate the relative pose between an image and a model, the input image needs to be represented as a point cloud. This is possible if depth information of the image and the intrinsic camera parameters are known. These requirements were also met for the \gls*{ASG} approach, which only estimates extrinsic camera parameters. In contrast to our approach, the ICP algorithm achieves an alignment of two point clouds. The relative camera pose between image and model can then be derived from the transformation required for the alignment.  

For the evaluation, we used 10 images which are registered once by the \gls*{ICP} algorithm and once by our \gls*{ASG} approach. Regarding the \gls*{ICP} method, the registration of an image is evaluated as successful if the reprojection error under the estimated camera pose falls below a threshold value of 20 pixels. Whether a successful registration was achieved by applying the \gls*{ASG} approach is still decided automatically. \cref{tab:ICP-success} shows the number of successfully registered test images for the different approaches. For a small average initial error (approx. 50 pixels), the ASG approach is superior to the \gls*{ICP} process. If the average initial error is almost 80 pixels, the success rate is identical with 6 out of 10 images. If the initial errors are further increased, the number of registered images remains constant for the \gls*{ICP} method, while it drops to 3 for the ASG approach.
\begin{table}
	\centering
	\begin{tabular}{p{0.13\textwidth}p{0.13\textwidth}p{0.13\textwidth}}
		\hline
		\textbf{Initial error} & \textbf{ASG} & \textbf{ICP} \\
		\hline \hline
		$30.25$& $9$ & $6$ \\
		$55.12$& $8$ & $6$ \\
		$80.22$& $6$ & $6$ \\ 
		$101.15$ & $3$ & $6$ \\	
		\hline	
	\end{tabular}
	\caption{This table shows the number of successfully registered images using \gls*{ASG} and \gls*{ICP} approaches. A total of 10 images were used for the test.}
	\label{tab:ICP-success}
\end{table}

\begin{table}
	\centering
	\begin{tabular}{llll}
		\hline	
		\textbf{Initial error} & \textbf{Test images} & \textbf{ASG} & \textbf{ICP} \\
		\hline \hline
		$30.25$& $6$ & $1.12 \pm 1.22$  & $11.16 \pm 4.45$  \\
		$55.12$& $6$ & $2.13 \pm 2.82$& $9.70 \pm 5.60$  \\
		$80.22$& $4$ & $1.89 \pm 1.25$& $10.41 \pm 4.50$  \\ 
		$90.51$& $3$ & $2.80 \pm 1.48$& $10.60 \pm 5.84$  \\
		\hline
	\end{tabular}
	\caption{This table indicates the accuracies achieved using \gls*{ICP} and \gls*{ASG} approaches. The mean value and the standard deviation of the achieved reprojection errors are given. For the error determination, only test images were used which were registered successfully by both procedures.}
	\label{tab:ICP}
\end{table}
\begin{figure*}
	\centering
	\resizebox{\textwidth}{!}{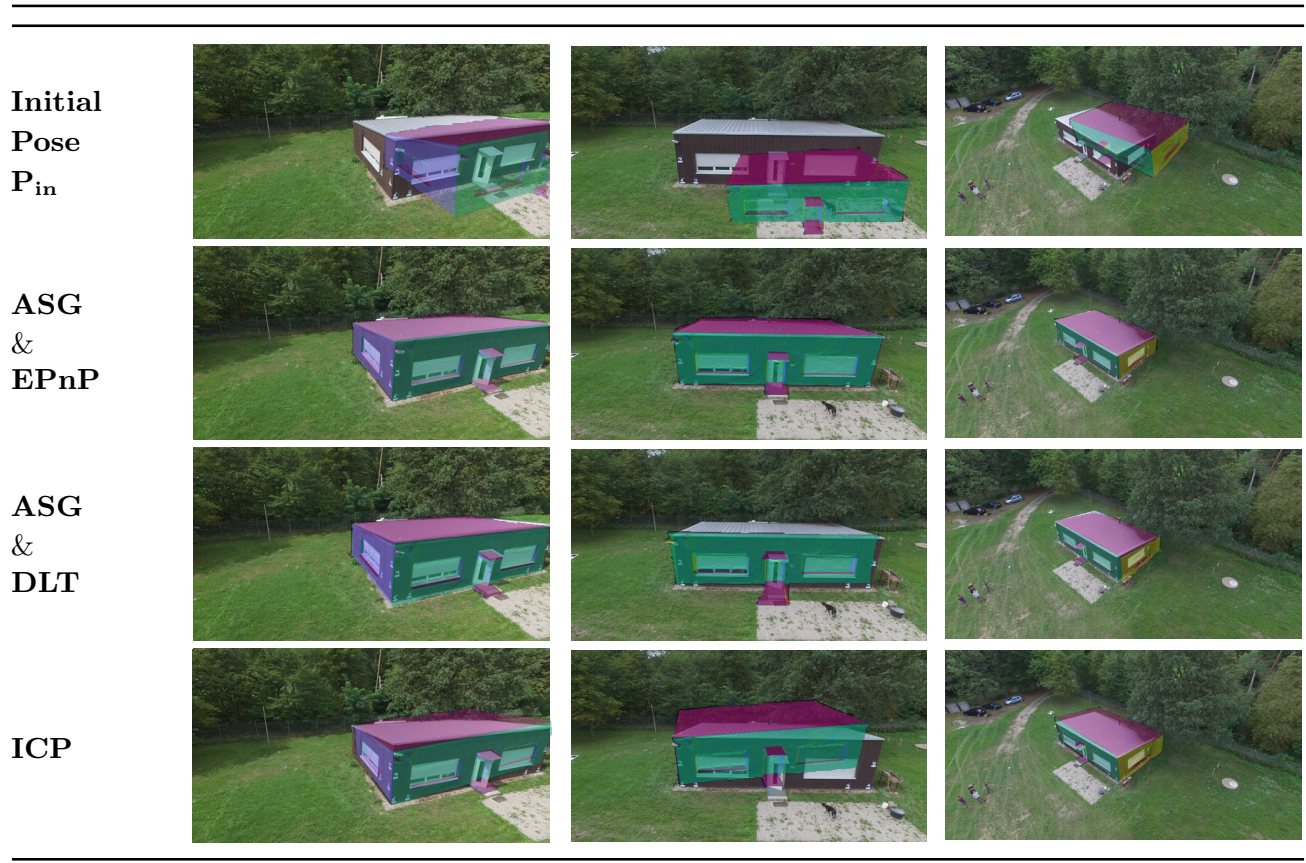}
	\caption{Qualitative results showing the input image together with a projection of the input model given the corresponding camera pose. %
		The input model of the building structure is projected in transparent colors, which encode the normal vectors of the model surfaces. %
		In the first row, the input model is projected given the initial camera pose. %
		These images clearly show the error in the camera pose with which our algorithm is initialized.
		The images in the second and third rows depict the results of our co-registration algorithm for known and unknown camera intrinsics. The fourth row shows the results of the ICP algorithm. %
		The results indicate that, in most cases, our algorithm achieves a good alignment between the input model and the input image. }
	\label{fig:comparison}
\end{figure*}

To compare the accuracy of the two methods, we determined the reprojection errors for all test images that can be successfully registered by both methods. \cref{tab:ICP} shows the corresponding error statistics (mean and standard deviation of reprojection errors) for increasing initial errors. It shows that the average error resulting from the pose estimates of the \gls*{ICP} method is about five times higher than that of the ASG method. Therefore, the ASG method allows to estimate camera poses closer to the true pose than the pose estimates provided by the \gls*{ICP} method. The poorer performance of the \gls*{ICP} method can partly be explained by noisy depth values of the aerial photographs. These have a negative effect on the accuracy of the point cloud, which results from the backprojection of the depth image. As mentioned before, the used point cloud of the model is also erroneous. The difficulty to adjust two noisy point clouds to each other is reflected in the high reprojection errors. ASG registration is more robust to errors in the input data. This can be explained by the fact that the geometric error minimized within the RANSAC scheme is based only on selected 2D-3D correspondences. These correspondences were previously determined using stable characteristics. In contrast, in the \gls*{ICP} procedure, all model points are included in the calculation of the error to be minimized. The observations of the comparison of registration procedures can be summarized as follows:
\begin{itemize}
	\item The accuracy of the estimated camera parameters using the ASG approach exceeds one of the ICP implementation. The reprojection error is on average five times smaller.
	\item At given initial poses with a mean reprojection error of less than 60 pixels, more images can be registered using the ASG approach than using the ICP implementation.
	\item For rough pose estimates with a mean reprojection error of over 80 pixels, more images can be registered by means of the ICP implementation than by means of the ASG approach.
\end{itemize}

\subsection{Qualitative Results}
\cref{fig:comparison} shows a selection of the results of the evaluated co-registration algorithms. It can be seen that our algorithm (second and third row) can handle large initial errors in the translation as well as minor errors in rotation. It can be observed that our proposed algorithm, based on the assumption of known intrinsic camera parameters (\gls*{ASG} with \gls*{EPnP}), gives the most accurate results. The estimation of the full camera pose (\gls*{ASG} with \gls*{DLT}) commonly yields similar results, but in some cases larger errors can be detected. For example, in the projection of the center image in \cref{fig:comparison}, the roof of the building disappears. Furthermore, it can be seen that registration by means of the \gls*{ICP} algorithm also improves the given poses $\mathrm{P}^{\mathrm{in}}$. However, the deviations from the true poses are greater than the deviations obtained using our algorithm.

\section{CONCLUSION \& FUTURE WORK}
\label{sec:conclusion}
\sloppy
In conclusion, we proposed an algorithm to automatically estimate the relative camera pose between aerial imagery and untextured 2.5D or 3D model data. To refine an initial guess of the camera pose, we compute feature-based dense correspondence fields between an aerial photograph and rendered images generated from different perspectives on the model. Since textural features are not present in the model, the compared features are derived from gradients that are based solely on the object geometry. To obtain such gradients related to the photograph as well as gradients related to the model, we use \gls*{ASG}, a method in which observable gradients from shadings are averaged over all possible light directions. Our evaluation shows that initial error-prone camera poses are significantly improved by our registration algorithm. Especially under the condition of calibrated camera sensors, good results are achieved. The error can thus be reduced to up to 1.07\percent of its initial value. With regard to the accuracy of estimated camera poses, our presented approach exceeds the results provided by the \gls*{ICP} algorithm. In addition, the automatic verification of the pose estimation provides a reliable statement about the success of the registration.

In our future work we want to integrate and evaluate further suitable methods for the estimation of poses from correspondences into our method instead of \gls*{DLT} or \gls*{EPnP}. In particular, approaches that include existing depth information of the input image are of interest. %
In addition, we want to evaluate the the difference in performance between methods that use line features, such as those extracted with the help of \gls*{ASG}, and methods that rely on point features.

{\footnotesize 
	\begin{spacing}{0.9}
		\bibliography{auto-co-registration} 
	\end{spacing}


\end{document}